\documentclass[letterpaper, 10 pt, conference]{ieeeconf}
\IEEEoverridecommandlockouts

\usepackage{bm}
\usepackage{amsmath} 
\usepackage{amssymb}  
\usepackage{amsfonts}

\usepackage{subfigure}
\usepackage{graphicx}
\usepackage{hyperref}

\usepackage{booktabs}
\usepackage{multirow}
\usepackage{tablefootnote}
\usepackage{threeparttable}

\title{\LARGE \bf
Segregator: Global Point Cloud Registration with Semantic and Geometric Cues
}

\author{Pengyu Yin$^{1}$, Shenghai Yuan$^{1}$, Haozhi Cao$^{1}$, Xingyu Ji$^{1}$, Shuyang Zhang$^{2}$, and Lihua Xie$^{1}$, \emph{Fellow, IEEE}
\thanks{$^{1}$Authors are with the Centre for Advanced Robotics Technology Innovation (CARTIN),  School of Electrical and Electronic Engineering, Nanyang Technological University, Singapore.
{\tt\small \{pengyu001, shyuan, haozhi002, xingyu001, elhxie\}@ntu.edu.sg}}%
\thanks{$^{2}$Authors are with the Department of Electronic and Computer Engineering, the Hong Kong University of Science and Technology, Clear Water Bay, Kowloon, Hong Kong SAR, China.
{\tt\small \{szhangcy\}@connect.ust.hk}}
\thanks{This research is supported by the National Research Foundation, Singapore under its Medium Sized Center for Advanced Robotics Technology Innovation.}
}

\begin{document}

\maketitle
\thispagestyle{empty}
\pagestyle{empty}

\begin{abstract}
This paper presents \emph{Segregator}, a global point cloud registration framework that exploits both semantic information and geometric distribution to efficiently build up outlier-robust correspondences and search for inliers. Current state-of-the-art algorithms rely on point features to set up putative correspondences and refine them by employing pair-wise distance consistency checks. However, such a scheme suffers from degenerate cases, where the descriptive capability of local point features downgrades, and unconstrained cases, where length-preserving (\textit{l}-TRIMs)-based checks cannot sufficiently constrain whether the current observation is consistent with others, resulting in a complexified NP-complete problem to solve. To tackle these problems, on the one hand, we propose a novel degeneracy-robust and efficient corresponding procedure consisting of both instance-level semantic clusters and geometric-level point features. On the other hand, Gaussian distribution-based translation and rotation invariant measurements (\textit{G}-TRIMs) are proposed to conduct the consistency check and further constrain the problem size. We validated our proposed algorithm on extensive real-world data-based experiments. The code is available: \url{https://github.com/Pamphlett/Segregator}.

\end{abstract}

\section{Introduction}
Point cloud registration finds the pose transformation between point clouds and is widely employed in many computer vision and robotics applications ranging from object recognition to robotic grasping to satisfy the demands for accurate pose estimation. In particular, it is an essential technique in the ego-motion estimation of mobile robots \cite{pomerleau2015review} to solve simultaneous localization and mapping (SLAM) problems, i.e., to estimate the pose transformation between either two consecutive LiDAR frames or more distant ones in a loop closing or relocalization scenario. 

\begin{figure}[htb]
    \centering
    \includegraphics[width=8.5cm]{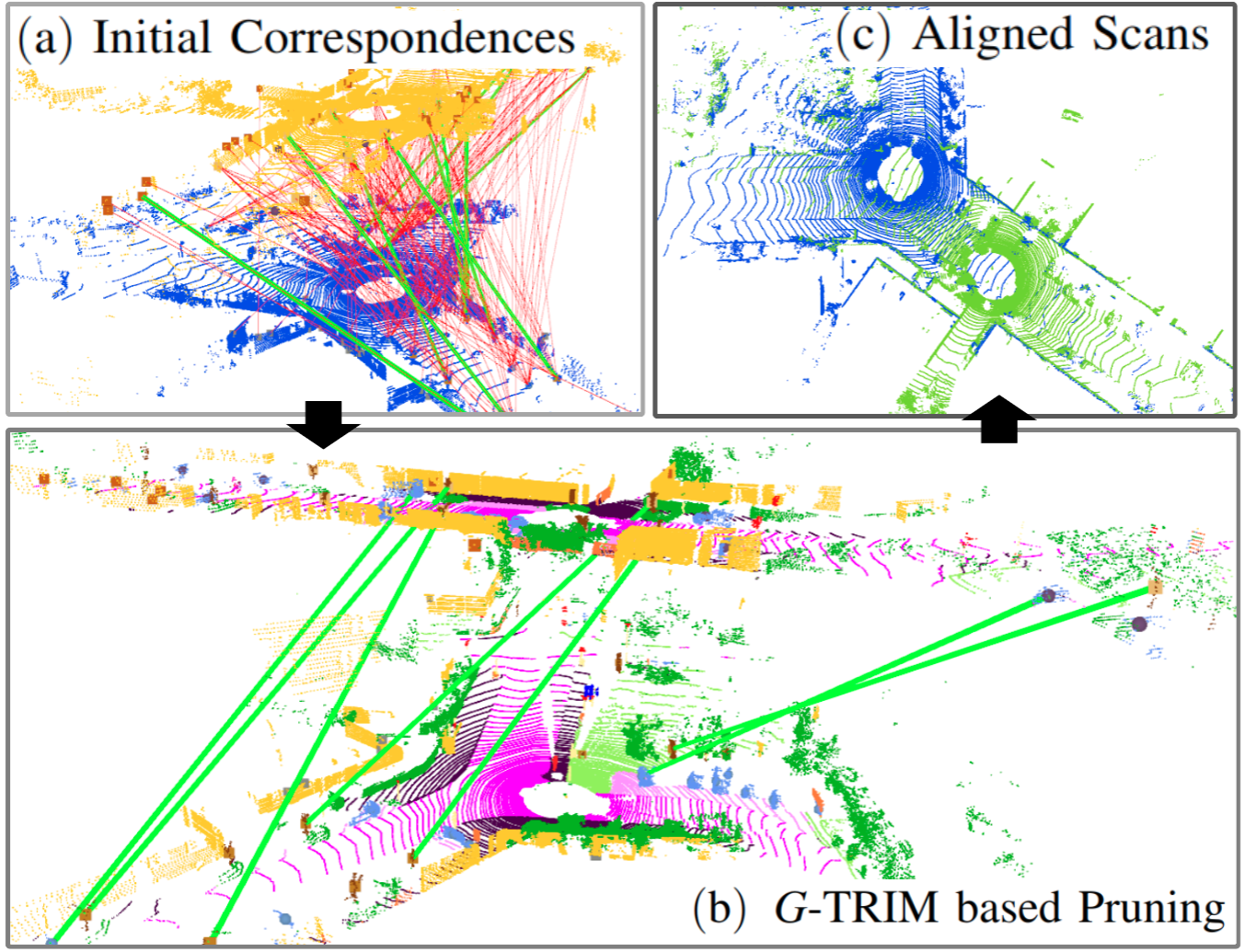}
    \caption{The registration pipeline of Segregator with two distant point clouds as inputs. (a) The initial source and target clouds are colored yellow and blue. We managed to set up outlier-robust correspondences leveraging both point features and semantic cues (section \ref{sec:corresponding}). (b) Inlier correspondences (green lines) are extracted by \textit{G}-TRIM based graph pruning (section \ref{Sec:G-TRIM}) from massive outliers (thinner red lines in (a)). (c) Final aligned point clouds with the transformed cloud being green.}
    \label{fig:method_intro}
    \vspace{-1em}
\end{figure}

Point cloud registration algorithms can be mainly divided into two categories, namely local and global registration, by whether an initial guess is needed. Local registration methods \cite{besl1992method} set up correspondences by a heuristic nearest neighbor (NN) search and generate satisfactory pose estimation results given a good initial guess. On top of traditional local registration algorithms \cite{besl1992method}, \cite{chetverikov2002trimmed}, a variety of robust estimation and probabilistic modeling techniques, such as M-estimators \cite{pomerleau2013comparing} and noise modeling \cite{segal2009generalized}, are introduced to boost local algorithms' performance and have shown promising results in many LiDAR odometry solutions \cite{pan2021mulls}, \cite{chen2022direct}. However, the performance of local algorithms still depends on a good initial value, which makes them prone to partial overlapping and the emergence of dynamic obstacles and outliers.

Unlike their local counterparts, global registration algorithms do not rely on any assumption of an initial guess, which is also the reason that there has been a growing interest in using them to solve loop closure and global (re)localization problems. However, these problems can not be tractably solved by traditional correspondence-free \cite{yang2015go} or correspondence-based \cite{zhou2016fast} global registration methods due to their high volumes of data points. More recently, correspondence-based methods leverage geometric primitives to set up putative correspondences to reduce problem size, and further prune built-up correspondences by employing robust estimation techniques \cite{lusk2021clipper}, \cite{9812018}, \cite{yang2020teaser}, \cite{shi2021robin}. Still, point feature only corresponding schemes suffer from degenerate cases \cite{9812018}; and using the single length preserving property cannot well constrain the consistency check problem. To sum up, there is a need to develop a degeneracy-robust correspondence establishing pipeline, as well as a more compact consistency checker for global registration problems in the autonomous driving scenario.

%


In this paper, we investigate the combination of semantic cues, obtained using neural networks \cite{milioto2019rangenet++}, \cite{tang2020searching}, with Maximum Inlier Clique (MIC), a robust estimation technique, to tackle the aforementioned problems. We make the following contributions:
\begin{enumerate}
    \item a semantic global point cloud registration framework, producing accurate pose estimations for low overlapping LiDAR scans in autonomous driving scenarios;
    \item a degeneracy-robust correspondence and consistency graph construction method consisting of both semantic clusters and geometric distributions (\textit{G}-TRIM), leading to an inlier correspondence searching algorithm with better efficiency and accuracy;
    \item extensive evaluations (over 80,000 pairs of scans) are conducted on publicly available data sets, proving superior performance of the proposed framework than current state-of-the-arts.
\end{enumerate}

\section{Related Work}

Early solutions for the point cloud registration problem adopt a framework iterating between the corresponding step and the error minimization step until certain termination criterion is satisfied \cite{chetverikov2002trimmed}, \cite{segal2009generalized}, \cite{serafin2015nicp}. A highly efficient but naive corresponding strategy, nearest neighbor (NN) search, is widely used to make the time complexity manageable. As stated by \cite{pomerleau2015review}, these algorithms can produce reasonable estimation results when the following assumption is satisfied: 
the portion of inlier correspondences acquired by NN search improves in each iteration. If that is not the case, algorithms could be trapped in local and often unsatisfying minima. Several commonly observed phenomena could contribute to the aforementioned convergence problem, including noisy/distant measurements, degenerate environments, and dynamic obstacles.

Branch-and-bound-based (BnB) global point cloud registration methods \cite{yang2015go}, \cite{bustos2017guaranteed} avoid the correspondence establishing process by enumerating the solution space and obtaining globally optimal solutions, whereas they run in exponential time, which prohibits them from being directly used in LiDAR scans with a large collection of points. Another category of global registration methods stays in the same framework as local ones, i.e., corresponding and error minimization, but being more computationally attractive by setting up correspondence only once \cite{choy2019fully}, \cite{briales2017convex}, \cite{yang2020teaser}. These deterministic correspondences are usually built via either learned \cite{choy2019fully} or traditional \cite{rusu2009fast} feature descriptors and then fed to an outlier pruning step to find inlier correspondences. 

In the correspondence setting up stage, however, degenerate cases could happen. As observed in \cite{9812018}, \cite{yang2020teaser}, point feature quality could downgrade as the size of the point cloud increases and thus those point feature-based correspondences could end up being very unreliable and contain less than two correct correspondences, making the rotation estimation impossible. Quatro \cite{9812018} alleviates the degenerate problem by restricting the degree of freedom of rotation to one and assuming point clouds to be co-planar in an urban scenario. Nevertheless, the quasi-SE(3) assumption could be challenged in non-flat areas, and the method does not fundamentally resolve the degenerate problem (e.g., in the correspondence establishing step). 

In terms of the outlier correspondence pruning stage, some local methods \cite{pan2021mulls}, \cite{zaganidis2018integrating} incorporate semantic masks to build better correspondences. A more conservative and robust way is to distinguish the inliers from the pre-built correspondences \cite{lusk2021clipper}, \cite{yang2020teaser}, \cite{bustos2017guaranteed}. GORE \cite{bustos2017guaranteed} exploits geometry features to avoid using branch and bound on the whole-size initial correspondence sets. Another solution is Teaser \cite{yang2020teaser}, which represents correspondences by a compatibility graph where vertexes are defined as the distance between points and edges are the results of the compatibility check. Then inliers are selected by finding the Maximum Inlier Clique (MIC) of the graph, which is in general NP-complete and requires exponential time to solve. Clipper~\cite{lusk2021clipper} further extends the formulation to a weighted scheme. However, all these methods are faced with scalability problems and do not fully exploit local geometric statistics, especially the distribution of points, resulting in searching for MIC in a bigger graph that could be computationally intensive.

\section{Methodology}
\subsection{Problem Formulation}
Given a pair of partially overlapping point clouds collected by the LiDAR sensor $\mathcal{P}=\left\{\mathbf{p}_i\right\}_{i=1}^{n}$ and $\mathcal{Q}=\left\{\mathbf{q}_j\right\}_{j=1}^{m}$ in two different coordinates, we seek to associate them into one common coordinate system. With unknown ground truth transformation, measurements $\mathbf{p}_i$ and $\mathbf{q}_j$ satisfy the following generative model

\begin{equation}
    \mathbf{p}_i=\mathbf{R}\mathbf{q}_j+\mathbf{t}+\mathbf{o}_{ij},
    \label{measurement fun}
\end{equation}
where index $i$ and $j$ denote one candidate in the raw correspondence sets $\mathcal{I}$. $\mathbf{T}$ denotes the ground truth transformation between the scans with 
\begin{equation*}
SO(3)\overset{\text{def}}{=}\left\{{\bf R}\in \mathbb{R}^{3\times3}:{\bf R}^\top{\bf R}={\bf{I}},\text{det}({\bf R})=1\right\}, \\
\end{equation*}
\begin{equation*}
SE(3)\overset{\text{def}}{=}\left\{\mathbf{T}=\begin{pmatrix} {\bf R} & {\bf t} \\ {\bf 0} & 1 \end{pmatrix}:{\bf R}\in SO(3), {\bf t}\in \mathbb{R}^3\right\}.
\end{equation*}

In Eq. (\ref{measurement fun}), $\mathbf{o}_{ij}$ can be modeled as a Gaussian distribution if $i,j\in\mathcal{I}_{GT}$ and is a random vector if not \cite{yang2020teaser}. Accordingly, the objective function of point cloud registration can be formulated as minimizing the measurement-wise error in given metric space,
\begin{equation}
    \hat{\mathbf{R}},\hat{\mathbf{t}}=\mathop{\arg\min}_{{\bf R}\in SO(3), {\bf t}\in \mathbb{R}^3}\sum\limits_{ij\in\mathcal{I}}\rho\left(e\left(\mathbf{p}_i-\mathbf{R}\mathbf{q}_j-\mathbf{t}\right)\right),
    \label{eq:ori objective}
\end{equation}
where $\rho(\cdot)$ and $e(\cdot)$ are the robust loss function and error metric respectively.

Accordingly, the point cloud registration problem can then be split into three steps: firstly, establish a set of measurement correspondences $\mathcal{I}\in[n]\times[m]:=\left\{1,\dots,n\right\}\times\left\{1,\dots,m\right\}$; secondly, obtain inlier correspondence sets; thirdly, minimize the residual error. The proposed algorithm will also come after this framework and be presented in detail in the following subsections.

\subsection{Degeneracy-Robust Correspondences Establishment} \label{sec:corresponding}
Although not explicitly stated in the literature, setting up correspondences suffers from a \textit{radical versus conservative} dilemma. Feature-based corresponding methods \cite{9812018}, \cite{choy2019fully} tend to be radical as they build one-to-one correspondence according to feature similarity. In large scenes (e.g., an urban driving scenario), the descriptive capability of local features downgrades due to either the sparse nature of LiDAR point clouds \cite{9812018} or the emergence of locally similar areas. Yet all-to-all correspondence $\mathcal{I}_C=[n]\times[m]$ ensures the full inclusion of the ground truth correspondence sets while being computationally expensive. Therefore, it is neither suitable to construct putative point-level correspondences radically nor to be too conservative to set up the all-to-all correspondence. To this end, we leverage semantic and point features to seek a balance between these two.

Given the source point cloud $\mathcal{P}=\left\{\mathbf{p}_i\right\}_{i=1}^{n}$ and a semantic assignment function $\lambda:\mathbb{R}^3\rightarrow\mathbb{N}$, where semantic labels are represented by natural numbers in $\mathbb{N}$. Each point $\mathbf{p}_i$ in the cloud could be enhanced by their per-point label $l\in\mathcal{L}$. The semantically enhanced point cloud could be represented as
\begin{equation}
    \mathcal{S}=\left\{s_i|s_i=\left\{\mathbf{p}_i,\lambda\left(\mathbf{p}_i\right)\right\},\forall\mathbf{p}_i\in\mathcal{P}\right\},
\end{equation}
where $s_i$ carries both its coordinate and semantic label.

We use the dynamic curved-voxel clustering (DCVC) \cite{zhou2021t} algorithm to segment point clouds for each label into disjoint sub-clouds. This will produce sets of geometrically approximate points with the same semantic label
\begin{equation}
\begin{aligned}
    \mathbb{C}^l=\{C_1,\dots,C_N|&C_k\subset\mathcal{P}, \\ &l=\lambda(\mathbf{p}_i)=\lambda(\mathbf{p}_j)\ \forall\mathbf{p}_i,\mathbf{p}_j\in C_k\}.
\end{aligned}
\end{equation}
For each cluster in $\mathbb{C}^l$, we compute its centroid $c$ and its corresponding covariance matrix $\Sigma$ as its geometric characteristics.
The procedure is repeated for points in the target point cloud $\mathcal{Q}=\left\{\mathbf{q}_j\right\}_{j=1}^{m}$ with the same semantic label $l$ and resulting in two semantic cluster sets $\mathbb{C}_\mathcal{P}^l$ and $\mathbb{C}_\mathcal{Q}^l$. Thereafter, all-to-all correspondence is established between these semantic clusters (SCs) with the label $l$
\begin{equation}
    \mathcal{I}_\text{SC}^l:=\left\{\left(C_N,C_M\right)\in\mathbb{C}_\mathcal{P}^l\times\mathbb{C}_\mathcal{Q}^l\right\}.
\end{equation}
Semantic correspondences are established for every semantic class with a meaningful definition of 'instance' $\mathcal{L}_\text{SC}\subset\mathcal{L}$ (e.g. car, trunk, traffic sign). And the final semantic cluster-level correspondence $\mathcal{I}_\text{SC}$ is generated by concatenating all correspondences for each semantic category
\begin{equation}
    \mathcal{I}_\text{SC}=\bigcup\mathcal{I}_\text{SC}^l\quad \forall l\in\mathcal{L}_\text{SC}.
\end{equation}
For other more environmental semantic categories $\mathcal{L}_\text{F}$ (e.g., building, vegetation), we still comply with traditional feature-based setup (e.g., FPFH in Quatro \cite{9812018}) to generate the other set of correspondences $\mathcal{I}_\text{F}$. This is because the euclidean clusters for these semantic classes are heavily viewpoint-dependent and, therefore, their geometric characteristics could be unstable. Accordingly, semantic labels are also used to eliminate cross-semantic class correspondences in $\mathcal{I}_\text{F}$. Finally, the final correspondence sets, $\mathcal{I}_\text{raw}$ is the combination of $\mathcal{I}_\text{SC}$ and $\mathcal{I}_\text{F}$.

The philosophy under the above-presented correspondence establishment procedure is threefold: Firstly, the semantic cluster is a combination of both cognitive and geometric information, which makes them a suitable choice for either high-level tasks (e.g., rule out coarse outliers by semantic label discrepancy) or low-level ones (e.g., geometric perception). Secondly, semantic cluster-based correspondences also alleviate degenerate cases as the all-to-all correspondence ensures full inclusion of the inlier correspondence sets. Thirdly, the presented formulation considers both semantic clusters and geometric features, which acts as belt and braces to ensure registration success.

One may argue that cluster centroids could also vary with viewpoint changes, and it would consequently affect the pose estimation precision as well as inlier correspondence selection. However, firstly, the proposed method can always act as a coarse alignment followed by a finer one (e.g., ICP); secondly, in the following section \ref{Sec:G-TRIM}, we demonstrate how our novel distribution-based (\textit{G}-TRIM) consistency check alleviates the unstable centroid problem by considering both positional and distribution information.

\subsection{\textit{G}-TRIM based Outlier Pruning}\label{Sec:G-TRIM}
With a set of noisy correspondences, $\mathcal{I}_\text{raw}$, the problem of outlier pruning is to find the largest inlier correspondence sets $\hat{\mathcal{I}}$, where all entries inside satisfy the following distance constraint under the optimal transformation ${\bf T}$
\begin{align}
\begin{split}
    \mathop{\max}_{\mathcal{I}\subset\mathcal{I}_\text{raw},\;\mathbf{T}\in \text{SE}\left(3\right)} \qquad\qquad &\left|\mathcal{I}\right| \\ 
    \text{s.t.} \left\|\mathbf{y}_i-\mathbf{Rx}_i-\mathbf{t}\right\|_2 \leq\epsilon,\ &\ \forall i\in \mathcal{I},
    \label{equ: inlier corres}
\end{split}
\vspace{-1em}
\end{align}
where $\mathbf{x}_i$ and $\mathbf{y}_i$ are the corresponding points, $\epsilon$ is the threshold of measurement error. 

Since the ground truth transformation $\mathbf{T}$ is unknown, one cannot leverage Eq. (\ref{equ: inlier corres}) to find inlier correspondences. Rather, a graph-theoretic solution that takes advantage of the length-preserving property is proposed in Teaser \cite{yang2020teaser}. To be more specific, given two pairs of points $\mathbf{a}_i$ and $\mathbf{a}_j$, $\mathbf{b}_i$ and $\mathbf{b}_j$, associated by the raw correspondence $\mathcal{I}_\text{raw}$. Length-based translation and rotation invariant measurements (\textit{l}-TRIMs) are calculated as $d_{ij}:=\textit{l}\text{-TRIM}_{\mathbf{a}_{ij}}/\textit{l}\text{-TRIM}_{\mathbf{b}_{ij}}=\left\|\mathbf{a}_i-\mathbf{a}_j\right\|_2/\left\|\mathbf{b}_i-\mathbf{b}_j\right\|_2$. Since we work in a rigid transformation scenario, $d_{ij}=1$ holds for no-noise cases. In real applications, a noisy measurement function is observed. Thus, two correspondences are mutually consistent with each other when the corresponding $d_{ij}$ is very close to 1. By computing \textit{l}-TRIMs for all possible correspondence pairs, an undirected graph $\mathcal{G}=\left(\mathcal{V},\mathcal{E}\right)$ where vertexes are correspondences and edges are built between mutually consistent correspondences (vertexes). Since our goal is about finding the largest correspondence inlier set, the maximum clique\footnote{Cliques of a graph refers to the complete subgraphs, and the maximum clique is a clique with
the most vertexes \cite{bron1973algorithm}} of $\mathcal{G}$ should be found, which is NP-complete and could typically be solved in exponential time. 

\begin{figure}
\vspace{0.5em}
\centering

\subfigure[\textit{l}-TRIM based consistency graph construction.]{\includegraphics[height=3cm]{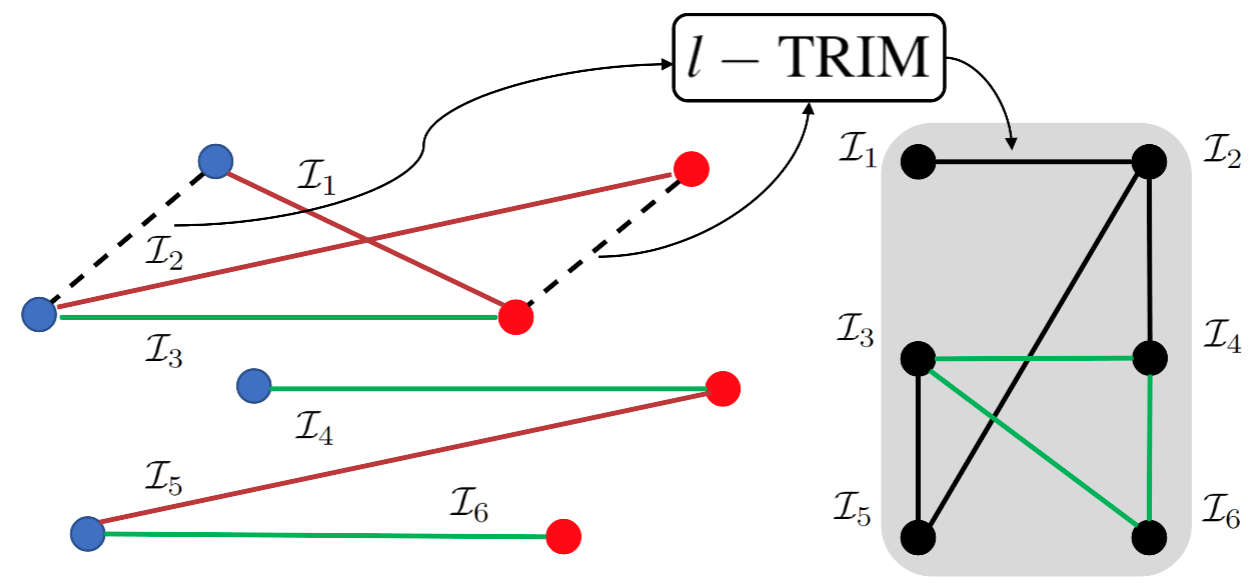} \label{lTRIM}}
\hspace{0.02in}
\subfigure[\textit{G}-TRIM based consistency graph construction.]{\includegraphics[height=3cm]{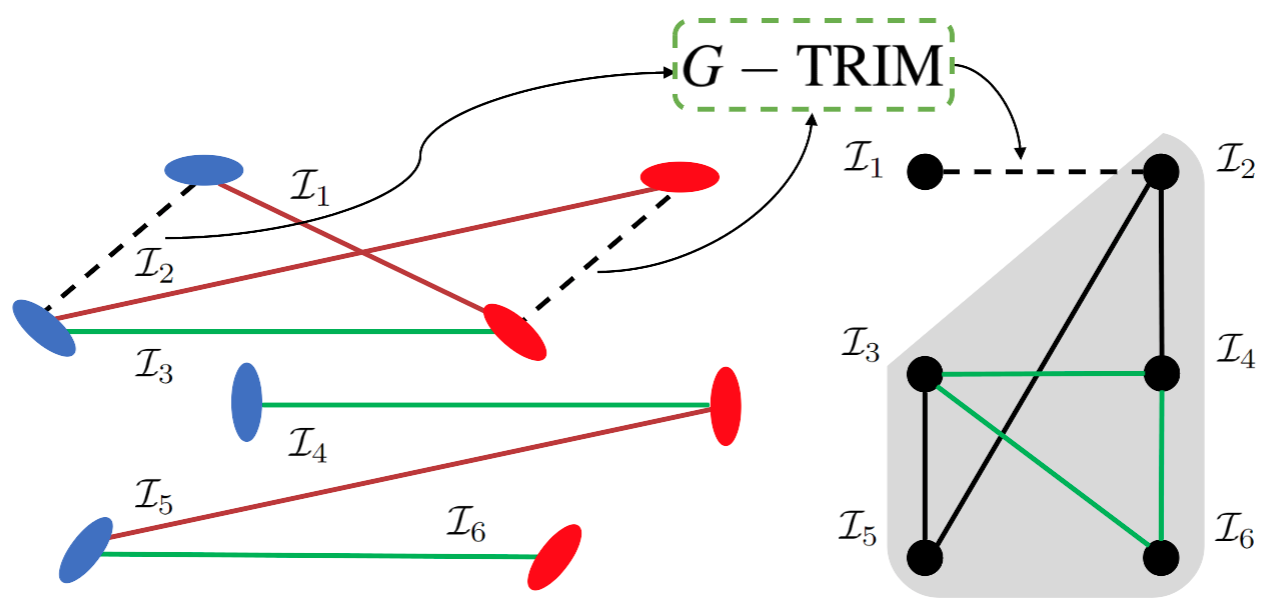} \label{GTRIM}}
\vspace{-0.5em}
\caption{Illustration of our proposed \textit{G}-TRIM-based consistency graph building method against the length-based (\textit{l}-TRIM) counterpart. In (b), \textit{G}-TRIM successfully resists the crossings outlier case ($\mathcal{I}_1$ and $\mathcal{I}_2$) by employing distributions (ellipsoids). Consistency graphs are shown in grey, and the final MIC is connected by green lines. \textit{G}-TRIM constructs a distribution-consistent graph with fewer vertexes.}
\label{Fig:lTRIM & GTRIM}
\setlength{\abovecaptionskip}{0.8cm}
\vspace{-1.5em}
\end{figure}

\textit{l}-TRIM-based consistency check retains correspondence pairs that satisfy the length-preserving property. However, it is not a reasonable constraint for all cases. As presented in Fig. \ref{lTRIM}, a pair of 4-point sets are shown in red and blue with a set of noisy correspondences ($\mathcal{I}_1$-$\mathcal{I}_6$ with outliers in red and inliers in green). We focus on the first two correspondences ($\mathcal{I}_1$ and $\mathcal{I}_2$) which are in a crossings situation. In this scenario, the \textit{l}-TRIM-based consistency check will treat them as inliers as swapping the corresponding sequence will not change the length. Accordingly, an edge between node $\mathcal{I}_1$ and $\mathcal{I}_2$ is constructed indicating their mutual consistency although they are indeed outlier correspondences.

We alleviate this problem by introducing a novel Gaussian distribution-based translation and rotation invariant measurement (\textit{G}-TRIM). In our formulation, the 3D position of every single point is modeled as a Gaussian distribution $\mathbf{p}_i\sim\mathcal{N}\left(\bm{\mu}_i,\bm{\Sigma}_i\right)$, where $\bm{\mu}_i$, the centroid, and $\bm{\Sigma}_i$, the covariance matrix, are calculated by $\bm{\mu}_i=\frac{1}{|\mathcal{V}_i|}\sum\mathbf{p}_i$ and $\bm{\Sigma}_i=\frac{1}{|\mathcal{V}_i|}\sum(\mathbf{p}_i-\bm{\mu}_i)^{\top}(\mathbf{p}_i-\bm{\mu}_i)$. $\mathcal{V}_i$ is a small patch of neighboring points around $\mathbf{p}_i$. The point-wise measurement $\mathbf{a}_{ij}$ also follows a Gaussian distribution $\mathbf{a}_{ij}:=\mathbf{a}_i-\mathbf{a}_j\sim\mathcal{N}\left(\mathbf{c}_i-\mathbf{c}_j,\bm{\Sigma}_{\mathbf{a}_i}+\bm{\Sigma}_{\mathbf{a}_j}\right)$, which we denote as Gaussian distribution-based translation and rotation invariant measurement (\textit{G}-TRIM). Consistency checks are conducted by computing the Wasserstein distance between these Gaussian distributions.
\begin{equation}
\begin{aligned}
    d^2\left(\mathbf{a}_{ij},\mathbf{b}_{ij}\right)&=W\left(\mathbf{a}_{ij},\mathbf{b}_{ij}\right) \\
    &=\left\|\mu_{\mathbf{a}_{ij}}-\mu_{\mathbf{b}_{ij}}\right\|_2^2 \\
    &+\text{Tr}\left(\bm{\Sigma}_{\mathbf{a}_{ij}}\!+\!\bm{\Sigma}_{\mathbf{b}_{ij}}\!-\!2\left(\bm{\Sigma}_{\mathbf{a}_{ij}}^{1/2}\bm{\Sigma}_{\mathbf{b}_{ij}}\bm{\Sigma}_{\mathbf{a}_{ij}}^{1/2}\right)^{1/2}\right).
\end{aligned}
\label{eq: wassers}
\end{equation}
We remark here that the first and second term in (\ref{eq: wassers}) captures positional and shape variations respectively.
Reasons for choosing the Wasserstein distance are twofold: First, the Wasserstein distance is a perfect distance metric for distributions, while some other distribution discrepancy measurements, like KL-divergence, are not in a symmetric form and can not be used as metric level measurements; Second, Wasserstein distance has an elegant closed-form solution, while measured entries are both Gaussian \cite{villani2009optimal}. 

It is worth noting that the above \textit{G}-TRIM based consistency check could successfully resist the crossings case (Fig. \ref{Fig:lTRIM & GTRIM}) as $W\left(\mathbf{a}_{ij},\mathbf{b}_{ij}\right)\neq W\left(\mathbf{a}_{ij},\mathbf{b}_{ji}\right)$ as we presented in Fig. \ref{GTRIM}, the formulation prevents the insertion of the edge between correspondences $\mathcal{I}_1$ and $\mathcal{I}_2$. Such crossings cases, however, could be commonly observed in $\mathcal{I}_{raw}$, especially for the all-to-all correspondence case. Moreover, \textit{G}-TRIM-based consistency check captures more geometric information compared to its length-based counterpart as it measures positional as well as the structural similarity between corresponding measurements. Considering the mixer of information could 1. partially solve the unstable centroid problem mentioned in \ref{sec:corresponding} as two corresponding centroids with tiny positional variation could have similar neighboring distribution and thus pass the consistency check; 2. serves in a belt and braces way with both length and distribution preserving property to solve the unconstrained problem.

Finding the inlier correspondence sets $\hat{\mathcal{I}}$ in Eq. (\ref{equ: inlier corres}) can be solved by incorporating the parallel maximum clique algorithm \cite{rossi2015parallel} to the consistency graph built by \textit{G}-TRIM.

\subsection{Pose Estimation}
With inlier correspondence sets $\hat{\mathcal{I}}$, we formulate the objective function in Eq. (\ref{eq:ori objective}) into the following truncated least squares (TLS) form to resist potential outliers further \cite{yang2020graduated}
\begin{equation}
    \hat{\mathbf{R}},\hat{\mathbf{t}}=\mathop{\arg\min}_{{\bf R}\in SO(3), {\bf t}\in \mathbb{R}^3}\sum\limits_{ij\in\hat{\mathcal{I}}}\min\left(\left\|\mathbf{p}_i-\mathbf{R}\mathbf{q}_j-\mathbf{t}\right\|_2,c_{ij}\right),
    \label{eq:final objective}
\end{equation}
with $c_{ij}$ the truncation parameter. Eq. (\ref{eq:final objective}) are then solved by leveraging Black-Rangarajan duality \cite{black1996unification} and graduated non-convexity (GNC) \cite{yang2020graduated}.

\section{Experimental Results}
In this section, we conduct comparative experiments to test the proposed algorithm against current state-of-the-arts. The proposed algorithm is implemented in C++. All experiments are conducted on a PC with an Intel Core 2.60GHz i5 CPU, 32Gb RAM, and Geforce RTX3060 GPU. We follow the parameter settings in \cite{zhou2021t} for clustering and empirically set the noise bound $c_{ij}$ in \ref{eq:final objective} to 0.2.

\textbf{Benchmark Data Set}
We choose the publicly available KITTI data set \cite{geiger2013vision} to conduct all experiments. Experiments are divided into two parts, namely the robustness test, and the loop closing test. Detailed experiment settings can be found in the following sections (\ref{sec: r test} and \ref{sec: lc test}) respectively.

\textbf{Baseline Methods}
Three state-of-the-arts are chosen to be baseline, consisting both local and global registration methods, namely V-GICP \cite{koide2021voxelized}, TEASER++ \cite{yang2020teaser} and Quatro \cite{9812018}. We leverage the open-sourced version of each algorithm. As TEASER++ \cite{yang2020teaser} was not intend to be a full registration method (more of a robust solver), we apply the FPFH-based correspondence built by Quatro \cite{9812018} to it. Moreover, all other methods employed OpenMP \cite{chandra2001parallel} in the MIC estimation, so we set the thread number to 4 accordingly when evaluating V-GICP \cite{koide2021voxelized} for a fair comparison.

\textbf{Error Metrics}
For all qualitative experiments, we adopt the relative pose error (RPE) \cite{pomerleau2013comparing} to evaluate the accuracy of the estimated pose $\hat{\mathbf{T}}$ against the ground truth $\mathbf{T}$. Note $\Delta\mathbf{T}=\mathbf{\hat{T}}\cdot\mathbf{T}^{-1}$ and $\Delta{x}$, $\Delta{y}$, $\Delta{z}$ are the positional entries in $\Delta\mathbf{T}$, then the translation error and rotation error are calculated by 
\begin{equation*}
    e_{trans}=\sqrt{\Delta{x}^2+\Delta{y}^2+\Delta{z}^2},
\end{equation*}
\begin{equation*}
    e_{rot}=\arccos{\left(trace(\Delta{{\bf T}})/2-1\right)}.
\end{equation*}

\begin{figure}[t]

\centering
\includegraphics[width=8.6cm]{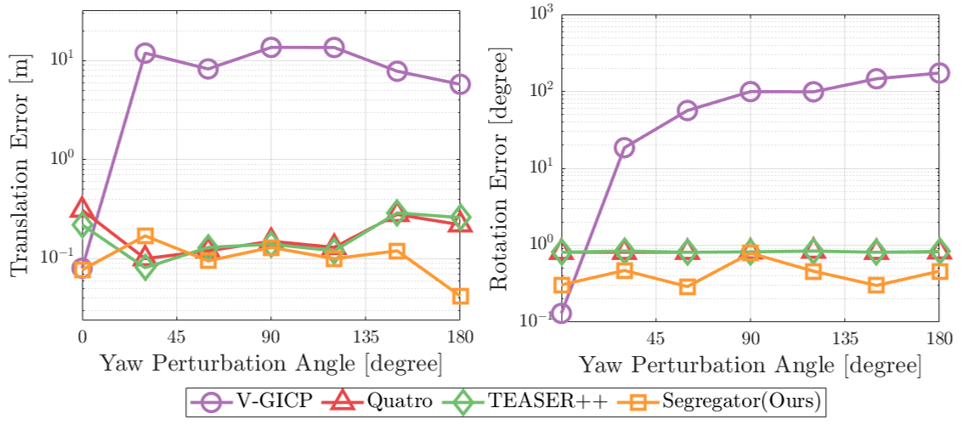}
\vspace{-1.5em}
\caption{Pose estimation with yaw angle perturbation.}
\label{fig:yaw angle res}
\vspace{-2em}
\end{figure}

\subsection{Robustness Test} \label{sec: r test}
The robustness factor is defined as the ability to perform normally under perturbations. For this purpose, we test whether algorithms are robust to translation, rotation, and also imperfect semantic masks. Accordingly, we manipulate the first and the 10th frame (approximately 5 meters apart with no rotational misalignment) of KITTI sequence 05 by adding different yaw angles perturbations as initial values and observe the performance of each algorithm. Results are shown in Fig. \ref{fig:yaw angle res}. V-GICP \cite{koide2021voxelized}, as the only local registration method, gets the most accurate pose estimation result in the case of no perturbation. However, its performance drastically collapsed when perturbation started to increase due to the naive NN-based correspondence establishing. All global methods are capable of generating reasonable pose estimations even under the most serious yaw perturbation, as all these methods contain a well-designed outlier-rejecting module. TEASER++ and Quatro behave very similarly due to that they are provided with the same FPFH-generated correspondences and have similar length-based inlier correspondence selection procedures. But it is also worth noting that the proposed algorithm, Segregator, is the only global registration method that can resist all yaw angle perturbations while also generating rather accurate pose estimation results as local state-of-the-art V-GICP. This is because our pose estimation measurements consist of both point-level features and instance-level semantic clusters, constituting a hierarchical measurement pool. The proposed mix-level correspondence setup (section \ref{sec:corresponding}) thus complements to each other and anchors each inlier measurement more precisely.

\begin{figure}[t]
\vspace{0.5em}
\centering
\includegraphics[width=8.5cm]{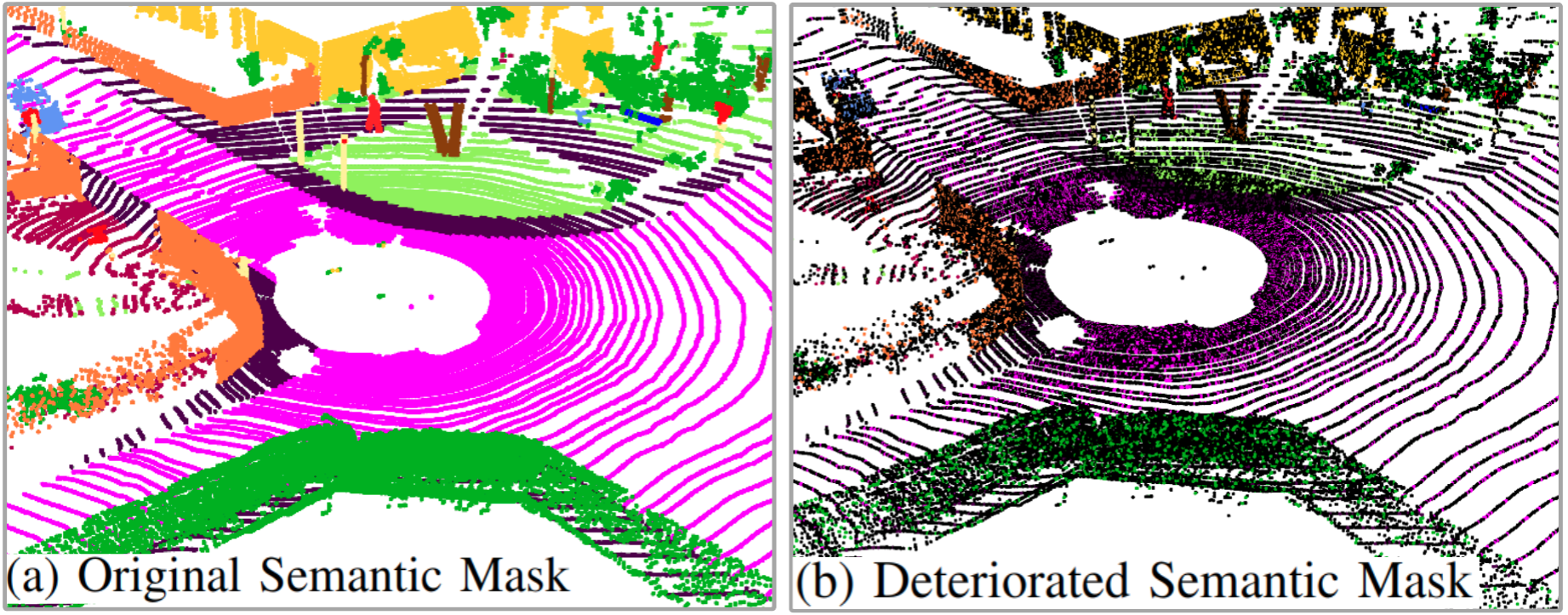}
\vspace{-0.5em}
\caption{Illustration of the original semantic mask inferred by SPVNAS \cite{tang2020searching} and the deteriorated one. We randomly set 90\% of the labels to unclassified (shown as black in (b)). Even with such a severely deteriorated semantic mask, Segregator could produce reasonable pose estimation results with a translation error of 0.27 m and a rotation error of 1.69 degrees due to the employment of both geometric and semantic features.}
\vspace{-1.5em}
\label{fig:semantic mask}
\end{figure}

Furthermore, we investigate the performance of Segregator against semantic label deterioration. Namely, how would the quality of the semantic mask affect the quality of pose estimation. We gradually increase the noise ratio in the semantic label predicted by SPVNAS \cite{tang2020searching} by randomly setting part of original labels as unclassified (marked black in Fig. \ref{fig:semantic mask}(b)). For each contamination level, we run our algorithm ten times and get the average translation and rotation error. As the results in Table \ref{table: semantic deter} have shown, the proposed method can provide satisfactory pose estimation result even under very challenging semantic mask deterioration with up to 90\% of noise. This is because semantic labels guide the formulation of object level representations, rather to be involved in a metric level computation. Thus Segregator won't fail as long as the semantic-guided clustering process don't fail.

\begin{table}[htb]
\caption{Registration Accuracy with Label Deterioration}
\begin{center}
\begin{tabular}{cccccc}
\toprule
Deterioration Rate (\%) & 10    & 30   & 50   & 70   & 90 \\
\midrule
$e_{trans}$ [m]   & 0.09  & 0.04 & 0.31 & 0.06 & 0.27 \\
$e_{rot}$ [$\deg$] &  0.20 & 0.22 & 0.21 & 1.70  & 1.70 \\
\bottomrule
\end{tabular}
\end{center}
\label{table: semantic deter}
\vspace{-2em}
\end{table}

\begin{table*}[t]
\vspace{0.4cm}
\setlength{\belowcaptionskip}{-0.2cm}
\caption{Success Rate Results on Kitti}
\vspace{-10pt}
\begin{center}
\scalebox{0.86}{
\begin{tabular}{cccccccccccccccc}
\toprule
\multirow{3}{*}[-0.7em]{Sequence} & \multicolumn{15}{c}{KITTI} \\
\cmidrule(lr){2-16}
 & \multicolumn{3}{c}{00} & \multicolumn{3}{c}{05} & \multicolumn{3}{c}{06} & \multicolumn{3}{c}{07} & \multicolumn{3}{c}{08}\\
 \cmidrule(lr){2-4}\cmidrule(lr){5-7}\cmidrule(lr){8-10}\cmidrule(lr){11-13}\cmidrule(lr){14-16}
\textit{Hardness} & \textit{easy} & \textit{medium} & \textit{hard} & \textit{easy} & \textit{medium} & \textit{hard} & \textit{easy} & \textit{medium} & \textit{hard} & \textit{easy} & \textit{medium} & \textit{hard} & \textit{easy} & \textit{medium} & \textit{hard} \\
\cmidrule(lr){1-16}
\textit{Num. of Pairs} & 8503 & 7527 & 21267 & 5155 & 5243 & 11600 & 1128 & 1113 & 2825 & 1311 & 1169 & 2877 & 2064 & 2750 & 9457 \\
\midrule
\text{V-GICP} \cite{koide2021voxelized}  & 95.6 & 34.1 & 14.9 & 82.6 & 34.6 & 25.3 & 94.6  & 66.7 & 35.1 & 80.0  & 36.5 & 0.8 & 0.24 & 0.0 & 4.9\\
\text{TEASER++} \cite{yang2020teaser} & 97.0 & 65.4 & 41.1 & 96.4 & 73.4 & 48.0 & 99.8  & 94.3 & 83.2 & 91.1 & 49.2 & 30.7 & 96.6 & 74.8 & 49.5 \\
\text{Quatro} \cite{9812018}  & 96.9 & 66.2 & 41.2 & 96.3 & 73.7 & 48.1 & 99.8  & 93.9 & 83.2 & 91.3 & 51.0 & 31.9 & 96.8 & 75.2 & 49.8\\
\text{Segator\dag (Ours)}& 98.6 & 88.3 & 75.4 & 77.8 & 72.2 & 59.0 & 91.0  & 74.6 & 63.8 & $\textbf{100.0}$ & 95.5 & 81.4 & 90.3 & 84.4 & 78.2\\
\text{Segregator (Ours)}    & $\textbf{99.7}$ & $\textbf{89.3}$ & $\textbf{80.7}$ & $\textbf{98.7}$ & $\textbf{93.2}$ & $\textbf{81.8}$ & $\textbf{100.0}$ & $\textbf{99.6}$ & $\textbf{98.3}$ & $\textbf{100.0}$ & $\textbf{98.1}$ & $\textbf{88.0}$ & $\textbf{96.9}$ & $\textbf{94.1}$ & $\textbf{88.8}$\\
\bottomrule
\end{tabular}
}
\end{center}
\begin{tablenotes}\footnotesize
\item \dag for ablation study.
\end{tablenotes}
\label{table:suc rate}
\vspace{-2.5em}
\end{table*}

\subsection{Protocol-based Loop Closing Test} \label{sec: lc test}

\begin{table}[t]
\caption{Statistics of Evaluation Data Set}
\begin{center}
\begin{tabular}{cccccc}
\toprule
\multirow{2}{*}[-0.3em]{Sequence} & \multicolumn{5}{c}{KITTI} \\
\cmidrule(lr){2-6}
& 00 & 05 & 06 & 07 & 08 \\
\midrule
\text{Num. of Scans}& 4541 & 2761 & 1101 & 1101 & 4071 \\
\text{Num. of Pairs}& 37297 & 21998 & 5066 & 5357 & 14271 \\
\text{Loop Direction}& \textit{Both} & \textit{Both} & \textit{Same} & \textit{Same} & $\textit{Reverse}$ \\
\text{\% Reverse loop}& 3\% & 5\% & 0\% & 0\% & 100\% \\
\bottomrule
\end{tabular}
\end{center}
\label{table:stats}
\vspace{-1.2em}
\end{table}

\begin{table}[h]
\caption{ATE, ARE and Timing on KITTI Sequence 06$^*$}
\begin{center}
\scalebox{0.85}{
\begin{tabular}{ccccc}
\toprule
\text{Method} & \text{Success Rate} & \text{APE [m/$^\circ$]} & \text{Total Time [s]} & \text{AIC} \\
\midrule
\text{V-GICP} & 55.4\% & $\textbf{0.07}$/$\textbf{0.10}$ & \textbf{0.09} & - \\
\text{TEASER++} & 89.0\% & 0.36/0.67 & 0.15 & 3025 \\
\text{Quatro} & 89.0\% & 0.36/0.66 & 0.15 &  3025 \\
\text{Segregator (Ours)} & $\textbf{99.0\%}$ & \underline{0.09}/0.27 & \underline{0.10} & 2687 \\
\text{Segregator-c2f (Ours)} & $\textbf{99.0\%}$ & $\textbf{0.07}$/\underline{0.12} & 0.15 & - \\
\bottomrule
\end{tabular}
}
\end{center}
\begin{tablenotes}\footnotesize
\item $^*$ Evaluations are conducted on successfully registered pairs only.
\end{tablenotes}
\label{table:timing}
\vspace{-2.5em}
\end{table}

In this section, algorithms are tested in a loop closure condition, which refers to registering two loop closure candidates. Loops add metric level constraints in the pose graph and reduce the cumulative error (a.k.a. odometry drift). However, building wrong loops could result in catastrophic failure. To have an all-round testing basis, we extract all possible loop candidates in several sequences in KITTI. This data is set up by running the following protocol through the ground truth pose $\mathbf{T}_k$ of each scan,
{\setlength{\abovedisplayskip}{0.2cm}
\begin{equation}
    \mathcal{X}_k=\left\{\mathbf{T}_i:r_1\leq\|\mathbf{t}_k-\mathbf{t}_i\|_2\leq r_2,|i-k|\ge m,\forall\mathbf{T}_i\in \mathcal{T}\right\},
    \label{equ:dataset proto}
\end{equation}
}
where $r_1$ and $r_2$ are predefined translation thresholds; $m$ is a loop closing parameter to rule out neighboring frames; $\mathcal{T}$ is the whole pose set. Thus all potential loop closure candidates are found by searching for all geometrically proximate scans but with a relatively large acquiring time offset. Statistics of resulted data set are presented in Table \ref{table:stats}. We include both unidirectional and reverse loops. In terms of quantity, the protocol-based loop closing data set contains 83,989 pairs of loop closure candidates. According to the translation discrepancy ($\|\mathbf{t}_k-\mathbf{t}_i\|_2$ in Eq. (\ref{equ:dataset proto})), we separate the whole data set into three categories, namely easy ($3-5$m), medium ($8-10$m) and hard ($10-15$m). The categories and measures are inspired by the recent Hilti SLAM challenge scoring scheme.
Apart from all other baseline methods, we also include a trimmed version of our proposed method, which is \text{Segator} in Table \ref{table:suc rate}. We replaced the presented \textit{G}-TRIM (\ref{Sec:G-TRIM}) based pruning with an ordinary length-based one (\textit{l}-TRIM) to have a clear view of its effectiveness. All registration results are considered to be successful when the translation error is smaller than $2$ meters and the rotation error is under $5^\circ$. We compute the ratio between successful registered frames and the number of whole pairs to be registered. Furthermore, the ground truth data for KITTI 08 \cite{geiger2013vision} is found to be relatively noisy so we use the pose data in Semantickitti \cite{behley2019semantickitti} estimated by Suma++ \cite{chen2019suma++}.

Matching success rate results for the aforementioned algorithms is shown in Table \ref{table:suc rate}. It is observed that our proposed method, Segregator, outperforms all other baselines by a margin especially on harder cases. While all algorithms are working comparably well on \textit{easy} cases, it is worth noting that even the state-of-the-art local registration method V-GICP \cite{koide2021voxelized}, as we mentioned earlier, can hardly deal with \textit{medium} and \textit{hard} situations for its naive correspondence setup. As for global baselines, TEASER++ \cite{yang2020teaser} and Quatro \cite{9812018} have very  similar overall performance with Quatro \cite{9812018} being slightly better. That is because these two algorithms are fed with exactly the same correspondence sets. Quatro \cite{9812018} employed a Quasi-SE(3) assumption and relaxed the number of correspondence that is needed to estimate the rotation to only one, which we thought is the reason for the performance boost. As for our ablation, Segator, we observe the overall performance falls with the trimmed \textit{G}-TRIM part thus shown validity of our proposition.

Furthermore, we evaluate the running time and average pose error (APE), of each algorithm on KITTI \cite{geiger2013vision} sequence 06 with results presented in Table \ref{table:timing}. Only successfully registered pairs are included in the pose error calculation for a meaningful comparison. V-GICP \cite{koide2021voxelized} achieved the least ATE and ARE as well as the total run-time. It is typical for local registration methods to outperform global ones in terms of accuracy as they employ more points in one scan and thus can anchor every point more precisely. Our original method, Segregator, achieves the best performance among all global registration methods by using semantic clusters and geometric distribution information to search for more geometrically corresponding measurements. We further install Segregator into a simple coarse to fine scheme (Segregator-c2f) where Iterative Closest Points (ICP) \cite{besl1992method} takes the estimation result from Segregator as an initial value. The simple integration obtains comparable precision compared to V-GICP \cite{koide2021voxelized} while being far more stable in terms of success rate. Moreover, Segregator is more computationally attractive than all other global registration methods by being faster in correspondence generation while also having a smaller size MIC problem to solve. Average inlier count (AIC) refers to the average amount of edges in the consistency graph. \textit{G}-TRIM, serves as a more constrained consistency checker, successfully ruling out distribution inconsistent and crossings cases \ref{Sec:G-TRIM}.

\section{CONCLUSIONS}
In this paper, we present a global point cloud registration algorithm leveraging both semantic and geometric information dubbed as Segregator. It is proved to be more robust in registering distant scans, which makes it a suitable choice in loop closure and relocalization scenario. In the future, we plan to extend Segregator with probabilistic modeling technique in the pose estimation part for more accurate estimations. 




\bibliographystyle{IEEEtran}
\bibliography{mybib}

\end{document}